\documentclass[10pt,twocolumn,letterpaper]{article}
\pdfoutput=1

\usepackage{cvpr}
\usepackage{color}
\usepackage{times}
\usepackage{epsfig}
\usepackage{graphicx}
\usepackage{amsmath}
\usepackage{amssymb}

\usepackage{booktabs}
\usepackage{pifont}
\usepackage{multirow}
\usepackage{placeins}
\usepackage{tabularx}

\usepackage{amsmath,amsfonts,bm}

\def\eqref#1{equation~\ref{#1}}

\def\1{\bm{1}}

\DeclareMathAlphabet{\mathsfit}{\encodingdefault}{\sfdefault}{m}{sl}
\SetMathAlphabet{\mathsfit}{bold}{\encodingdefault}{\sfdefault}{bx}{n}

\newcommand{\cmark}{\ding{51}}%
\newcommand{\xmark}{\ding{55}}%

\usepackage{algorithm}
\usepackage{algpseudocode}
\usepackage{setspace}
\usepackage{amsmath}

\usepackage[breaklinks=true,bookmarks=false]{hyperref}

\cvprfinalcopy 

\def\cvprPaperID{****} 

\begin{document}

\title{MetaFuse: A Pre-trained Fusion Model for Human Pose Estimation}

\author{Rongchang Xie$^{1,6}$, ~Chunyu Wang$^{5}$, ~Yizhou Wang$^{2,3,4}$\\
	\normalsize $^{1}$Center for Data Science, Peking University  ~~~~~~ $^{2}$ Adv. Inst. of Info. Tech., Peking University  \\
	\normalsize $^{3}$Center on Frontiers of Computing Studies, Peking University ~~ $^{4}$CS Dept., Peking University\\
	\normalsize $^{5}$Microsoft Research Asia  ~~~~~  $^{6}$Deepwise AI Lab \\
	{\tt\small \{rongchangxie, yizhou.wang\}@pku.edu.cn, chnuwa@microsoft.com}
}
\maketitle


\begin{abstract}
Cross view feature fusion is the key to address the occlusion problem in human pose estimation. The current fusion methods need to train a separate model for every pair of cameras making them difficult to scale. In this work, we introduce \emph{MetaFuse}, a pre-trained fusion model learned from a large number of cameras in the Panoptic dataset. The model can be efficiently adapted or finetuned for a new pair of cameras using a small number of labeled images. The strong adaptation power of MetaFuse is due in large part to the proposed factorization of the original fusion model into two parts--- (1) a generic fusion model shared by all cameras, and (2) lightweight camera-dependent transformations. Furthermore, the generic model is learned from many cameras by a meta-learning style algorithm to maximize its adaptation capability to various camera poses. We observe in experiments that \emph{MetaFuse} finetuned on the public datasets outperforms the state-of-the-arts by a large margin which validates its value in practice.
\end{abstract}

\begin{figure}
	\centering
	\includegraphics[width=3.25in]{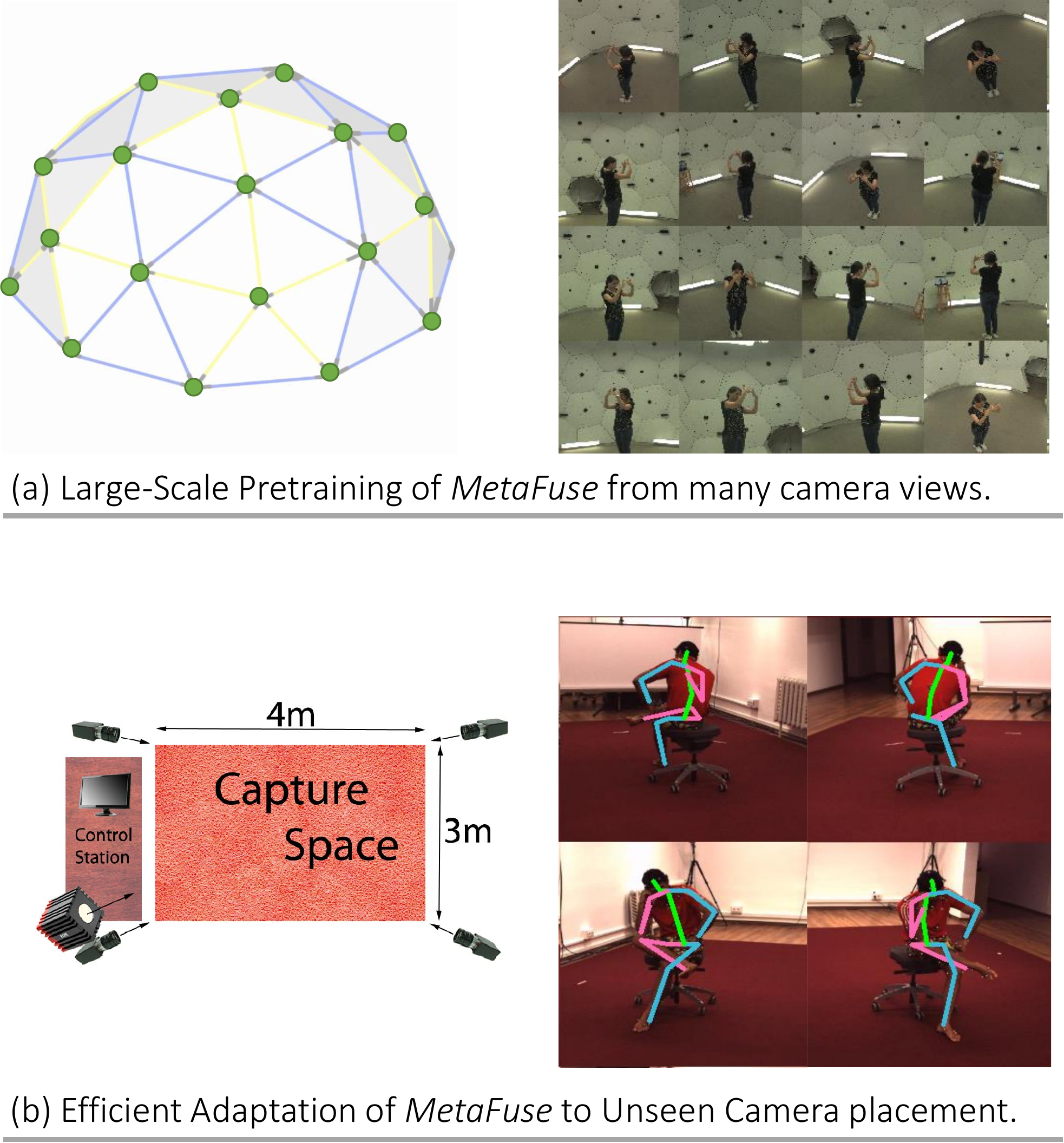}
	\caption{Concept of \emph{MetaFuse}. We learn a pre-trained feature fusion model from a large number of cameras, \ie the green dots in (a). Then for a new environment, we finetune the pre-trained model for each camera pair using only a few training data to get a customized $2$D pose estimator. The feature fusion allows us to localize the $2$D joints even when occlusion occurs as in (b).}
	\label{fig:sketch}
\end{figure}

\section{Introduction}
Estimating $3$D human pose from multi-view images has been a longstanding goal in computer vision. Most works follow the pipeline of first estimating $2$D poses in each camera view and then lifting them to $3$D space, for example, by triangulation \cite{iskakov2019learnable} or by pictorial structure model \cite{qiu2019cross}. However, the latter step generally depends on the quality of $2$D poses which unfortunately may have large errors in practice especially when occlusion occurs.

Multi-view feature fusion \cite{jafarian2018monet,qiu2019cross} has great potential to solve the occlusion problem because a joint occluded in one view could be visible in other views. The most challenging problem in multi-view fusion is to find the corresponding locations between different cameras. In a recent work \cite{qiu2019cross} , this is successfully solved by learning a fusion network for each pair of cameras (referred to as \emph{NaiveFuse} in this paper). However, the learned correspondence is dependent on camera poses so they need to retrain the model when camera poses change which is not flexible.

This work aims to address the flexibility issue in multi-view fusion. To that end, we introduce a pre-trained cross view fusion model \emph{MetaFuse}, which is learned from a large number of camera pairs in the CMU Panoptic dataset \cite{joo2015panoptic}. The fusion strategies and learning methods allow it to be rapidly adapted to unknown camera poses with only a few labeled training data. See Figure \ref{fig:sketch} for illustration of the concept. One of the core steps in \emph{MetaFuse} is to factorize \emph{NaiveFuse} \cite{qiu2019cross} into two parts: a generic fusion model shared by all cameras and a number of lightweight affine transformations. We learn the generic fusion model to maximize its adaptation performance to various camera poses by a meta-learning style algorithm. In the testing stage, for each new pair of cameras, only the lightweight affine transformations are finetuned utilizing a small number of training images from the target domain.

We evaluate \emph{MetaFuse} on three public datasets including H36M \cite{ionescu2014human3}, Total Capture \cite{trumble2017total} and CMU Panoptic \cite{joo2015panoptic}. The pre-training is only performed on the Panoptic dataset which consists of thousands of camera pairs. Then we finetune \emph{MetaFuse} on each of the three target datasets to get customized $2$D pose estimators and report results. For example, on the H36M dataset, \emph{MetaFuse} notably outperforms \emph{NaiveFuse} \cite{qiu2019cross} when $50$, $100$, $200$ and $500$ images are used for training the fusion networks, respectively. This validates the strong adaptation power of \emph{MetaFuse}. In addition, we find that \emph{MetaFuse} finetuned on $50$ images \footnote{Labeling human poses for $50$ images generally takes several minutes which is practical in many cases.} already outperforms the baseline without fusion by a large margin. For example, the joint detection rate for elbow improves from $83.7\%$ to $86.3\%$.

We also conduct experiments on the downstream $3$D pose estimation task. On the H36M dataset, \emph{MetaFuse} gets a notably smaller $3D$ pose error than the state-of-the-art. It also gets the smallest error of $32.4$mm on the Total Capture dataset. It is worth noting that in those experiments, our approach actually uses significantly fewer training images from the target domain compared to most of the state-of-the-arts. The results validate the strong adaptation capability of \emph{MetaFuse}.

\subsection{Overview of MetaFuse}
\begin{figure}
	\centering
	\includegraphics[width=3.2in]{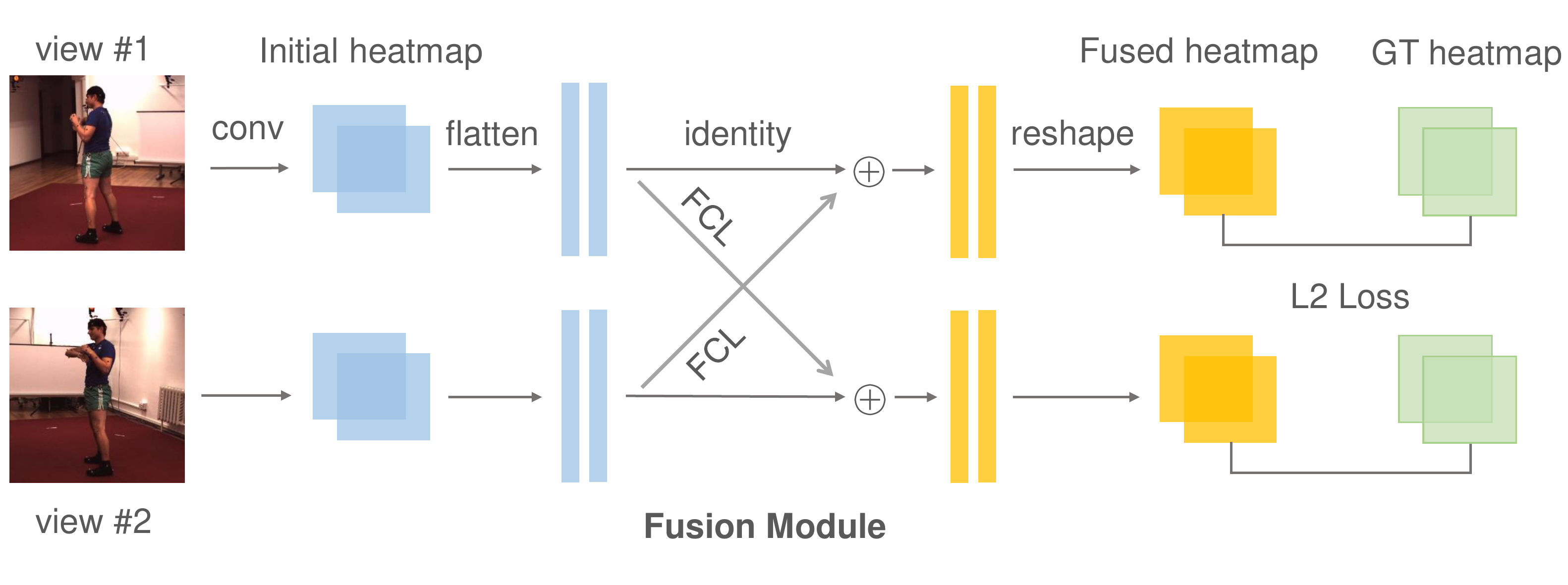}
	\caption{The \emph{NaiveFuse} model. It jointly takes two-view images as input and outputs $2$D poses simultaneously for both views in a single CNN. The fusion module consists of multiple FCLs with each connecting an ordered pair of views. The weights, which encode the camera poses, are learned end-to-end from data.}
	\label{fig:pipeline}
\end{figure}

\emph{NaiveFuse} learns the spatial correspondence between a pair of cameras in a supervised way as shown in Figure \ref{fig:pipeline}. It uses a Fully Connected Layer (FCL) to densely connect the features at different locations in the two views. A weight in FCL, which connects two features (spatial locations) in two views, represents the probability that they correspond to the same $3$D point. The weights are learned end-to-end together with the pose estimation network. See Section \ref{sec:naivefusion} for more details. One main drawback of \emph{NaiveFuse} is that it has many parameters which requires to label a large number of training data for \emph{every} pair of cameras. This severely limits its applications in practice.

To resolve this problem, we investigate how features in different views are related geometrically as shown in Figure \ref{fig:epipolar_geometry}. We discover that \emph{NaiveFuse} can be factorized into two parts: a \emph{generic} fusion model shared by all cameras as well as a number of \emph{camera-specific} affine transformations which have only a few learnable parameters (see Section \ref{sec:metafuse}). In addition, inspired by the success of meta-learning in the few-shot learning literature \cite{finn2017model,lee2018gradient,santoro2016meta}, we propose a \emph{meta-learning} algorithm to learn the generic model on a large number of cameras to maximize its adaptation performance (see Section \ref{sec:metalearning}). This approach has practical values in that, given a completely new multi-camera environment and a small number of labeled images, it can significantly boost the pose estimation accuracy.

\section{Related work}
\paragraph{Multi-view Pose Estimation}
We classify multi-view $3$D pose estimators into two classes. The first class is model based approaches such as \cite{liu2011markerless,bo2010twin,gall2010optimization,rhodin2018learning}. They define a body model as simple primitives such as cylinders, and optimize their parameters such that the model projection matches the image features. The main challenge is the difficult non-linear non-convex optimization problems which has limited their performance to some extent.

With the development of $2$D pose estimation techniques, some approaches such as  \cite{amin2013multi,Chen_2019_CVPR,burenius20133D,PavlakosZDD17,belagiannis20143d,dong2019fast,qiu2019cross} adopt a simple two-step framework. They first estimate $2$D poses from multi-view images. Then with the aid of camera parameters (assumed known), they recover the corresponding $3$D pose by either triangulation or by pictorial structure models. For example in \cite{amin2013multi}, the authors obtain $3$D poses by direct triangulation. Later the authors in \cite{burenius20133D} and in \cite{PavlakosZDD17} propose to apply a multi-view pictorial structure model to recover $3$D poses. This type of approaches have achieved the state-of-the-art performance in recent years. 

Some previous works such as \cite{amin2013multi,jafarian2018monet,qiu2019cross} have explored multi-view geometry for improving $2$D human pose estimation. For example, Amin \etal \cite{amin2013multi} propose to jointly estimate $2$D poses from multi-view images by exploring multi-view consistency. It differs from our work in that it does not actually \emph{fuse} features from other views to obtain better $2$D heatmaps. Instead, they use the multi-view $3$D geometric relation to \emph{select} the joint locations from the  ``imperfect" heatmaps. In \cite{jafarian2018monet}, multi-view consistency is used as a source of supervision to train the pose estimation network which does not explore multi-view feature fusion. \emph{NaiveFuse} \cite{qiu2019cross} is proposed for the situation where we have sufficient labeled images for the target environment. However, it does not work in a more practical scenario where we can only label a few images for every target camera. To our knowledge, no previous work has attempted to solve the multi-view fusion problem in the context of few-shot learning which has practical values.

\paragraph{Meta Learning} Meta-learning refers to the framework which uses one learning system to optimize another learning system \cite{vilalta2002perspective}. It learns from task distributions rather than a single task \cite{ravi2016optimization,santoro2016meta} with the target of rapid adaptation to new tasks. It has been widely used in few-shot classification \cite{lee2018gradient,santoro2016meta,snell2017prototypical} and reinforcement learning \cite{duan2016rl,mishra2018a} tasks. Meta learning can be used as an optimizer. For example, Andrychowicz \etal \cite{andrychowicz2016learning} use LSTM meta-learner to learn updating base-learner, which outperforms hand-designed optimizers on the training tasks. For classification, Finn \etal \cite{finn2017model} propose Model-Agnostic Meta-Learning (MAML) to learn good parameter initializations which can be rapidly finetuned for new classification tasks. Sun \etal \cite{Sun_2019_CVPR} propose Meta-Transfer learning that learns scaling and shifting functions of DNN weights to prevent catastrophic forgetting. The proposed use of meta-learning to solve the adaptation problem in cross view fusion has not been studied previously, and has practical values.

\section{Preliminary for Multi-view Fusion}
\label{sec:naivefusion}
\begin{figure}
	\centering
	\includegraphics[width=0.95\linewidth]{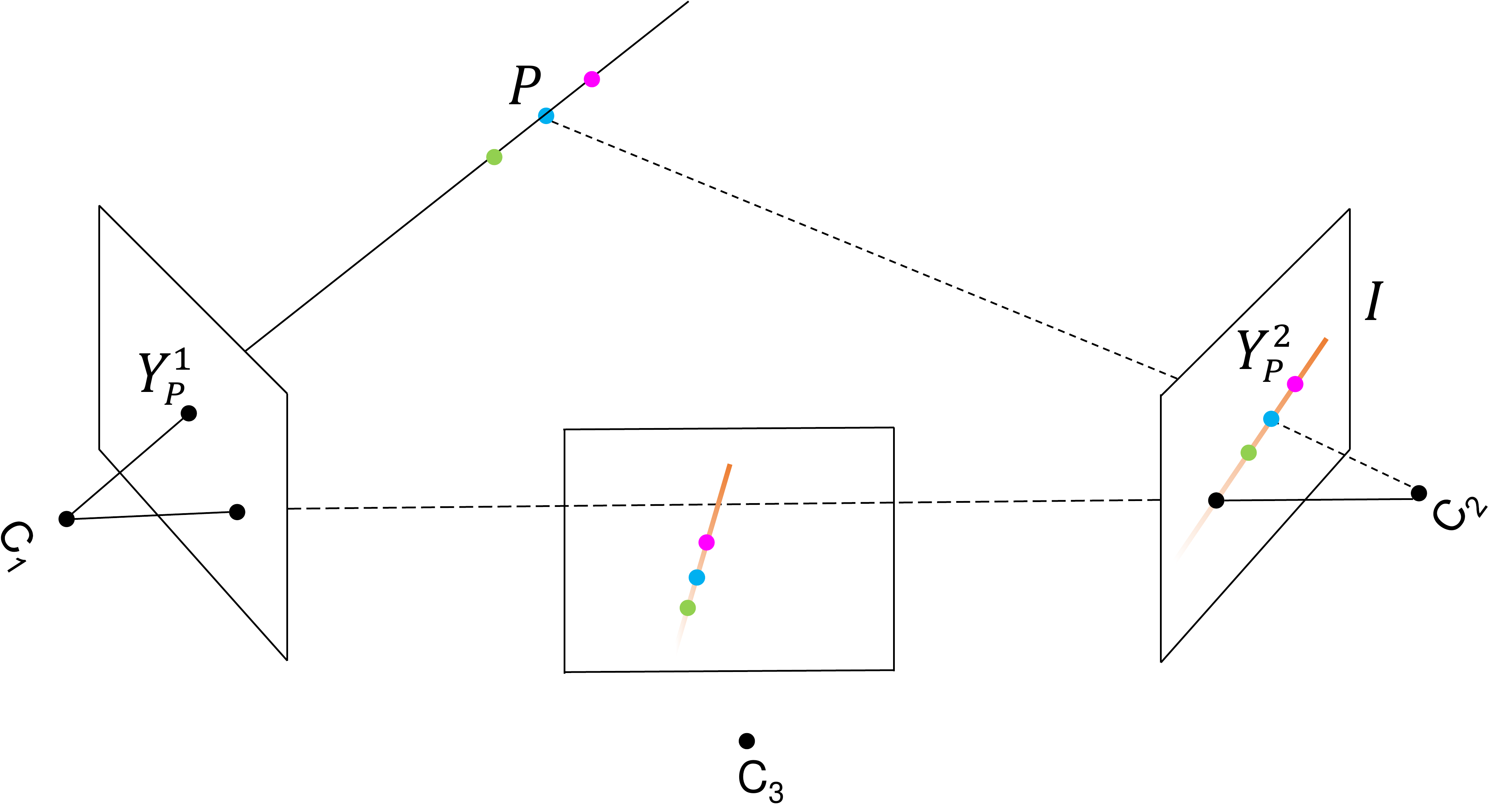}
	\caption{Geometric illustration of multi-view feature fusion. An image point $Y_P^1$ back-projects to a ray in 3D defined by the first camera center $C_1$ and $Y_P^1$. This line is imaged as $I$ in the second view. The 3D point $P$ which projects to $Y_P^1$ must lie on this ray, so the image of $P$ must lie on $I$. If the camera poses change, for example, we move the camera $2$ to $3$, then we can approximately get the corresponding line by applying an appropriate affine transformation to $I$. See section \ref{sec:metafuse}. }
	\label{fig:epipolar_geometry}
\end{figure}

We first present the basics for multi-view feature fusion \cite{hartley2003multiple,jafarian2018monet,qiu2019cross} to lay the groundwork for \emph{MetaFuse}. Let $\bm{P}$ be a point in $3$D space as shown in Figure  \ref{fig:epipolar_geometry}. The projected $2$D points in view $1$ and $2$ are $\bm{Y}_P^1 \in \mathcal{Z}_1$ and $\bm{Y}_P^2 \in \mathcal{Z}_2$, respectively. The $\mathcal{Z}_1$ and $\mathcal{Z}_2$ denote the set of pixel coordinates in two views, respectively. The features of view $1$ and $2$ at different locations are denoted as $\mathcal{F}^1=\{\bm{x}_1^1, \cdots, \bm{x}_{|\mathcal{Z}_1|}^1\}$ and $\mathcal{F}^2=\{\bm{x}_1^2, \cdots, \bm{x}_{|\mathcal{Z}_2|}^2\}$. The core for fusing a feature $\bm{x}_i^1$ in view one with those in view two is to establish the correspondence between the two views:
\begin{equation}
\bm{x}_i^1 \leftarrow \bm{x}_i^1 + \sum_{j=1}^{|\mathcal{Z}_2|}{\omega_{j,i} \cdot \bm{x}_j^2}, \quad \forall i \in \mathcal{Z}_1,
\label{eq:update_rule}
\end{equation}
where $\omega_{j,i}$ is a scalar representing their correspondence relation--- $\omega_{j,i}$ is positive when $\bm{x}_i^1$ and $\bm{x}_j^2$ correspond to the same $3$D point. It is zero when they correspond to different $3$D points. The most challenging task is to determine the values of all $\omega_{j,i}$ for each pair of cameras (\ie to find the corresponding points).

\paragraph{Discussion}
For each point $\bm{Y}_P^1$ in view $1$, we know the corresponding point $\bm{Y}_P^2$ has to lie on the epipolar line $I$. But we cannot determine the exact location of $\bm{Y}_P^2$ on $I$. Instead of trying to find the exact pixel to pixel correspondence, we fuse $\bm{x}_i^1$ with all features on line $I$. Since fusion happens in the \emph{heatmap layer}, ideally, $\bm{x}_j^2$ has large values near $\bm{Y}_P^2$ and zeros at other locations on the epipolar line $I$. It means the non-corresponding locations on the line will not contribute to the fusion. So fusing all pixels on the epipolar line is an appropriate solution. 

\paragraph{Implementation}
The above fusion strategy is implemented by FCLs (which are appended to the pose estimation network) in \emph{NaiveFuse} as shown in Figure \ref{fig:pipeline}. The whole network, together with the FCL parameters, can be trained end-to-end by enforcing supervision on the fused heatmaps. 
However, FCL naively connects each pixel in one view with all pixels in the other view, whose parameters are position-sensitive and may undergo dramatic changes even when the camera poses change slightly. So it is almost impossible to learn a pre-trained model that can be adapted to various camera poses using small data as our MetaFuse. In addition, the large FCL parameters increase the risk of over-fitting to small datasets and harm its generalization ability.

Note that we do not claim novelty for this \emph{NaiveFuse} approach as similar ideas have been explored previously such as in \cite{jafarian2018monet,qiu2019cross}. Our contributions are two-fold. First, it reformulates \emph{NaiveFuse} by factorizing it into two smaller models which significantly reduces the number of learnable parameters for each pair of cameras in deployment. Second, we present a meta-learning style algorithm to learn the reformulated fusion model such that it can be rapidly adapted to unknown camera poses with small data.

\section{MetaFuse}
\label{sec:metafuse}
Let $\bm{\omega^{\text{base}}} \in \mathcal{R}^{H \times W}$ be a \emph{basic} fusion model, \ie the fusion weight matrix discussed in Section \ref{sec:naivefusion}, which connects ONE pixel in the first view with all $H \times W$ pixels in the second view. See Figure \ref{fig:epipolar_geometry} for illustration. For other pixels in the first view, We will construct the corresponding fusion weight matrices by applying appropriate affine transformations to the basic weight matrix $\bm{\omega^{\text{base}}}$. In addition, we also similarly transform $\bm{\omega^{\text{base}}}$ to obtain customized fusion matrices for different camera pairs. In summary, this basic fusion weight matrix (\ie the \emph{generic} model we mentioned previously) is shared by all cameras. We will explain this in detail in the following sections.

\subsection{Geometric Interpretation}
\label{sec:affinefusion}
From Figure \ref{fig:epipolar_geometry}, we know $\bm{Y}_P^1$ corresponds to the line $I$ in camera $2$ which is characterized by $\bm{\omega^{\text{base}}}$. If camera $2$ changes to $3$, we can obtain the epipolar line by applying an appropriate affine transformation to $I$. This is equivalent to applying the transformation to $\bm{\omega^{\text{base}}}$. Similarly, we can also adapt $\bm{\omega^{\text{base}}}$ for different pixels in view one. Let $\bm{\omega_{i}} \in \mathcal{R}^{H \times W}$ be the fusion model connecting the $i_{th}$ pixel in view 1 with all pixels in view 2. We can compute the corresponding fusion model by applying a dedicated transformation to $\bm{\omega^{\text{base}}}$
\newcommand\Tau{\mathrm{T}}
\begin{equation}
\bm{\omega}_{i} \leftarrow \mathcal{\Tau}^{\theta_i}(\bm{\omega^{\text{base}}}), \quad \forall i,
\label{eq:affine_rule}
\end{equation}
where $\Tau$ is the affine transformation and $\theta_i$ is a six-dimensional affine transformation parameter for the $i_{th}$ pixel which can be learned from data. See Figure \ref{fig:base_weight} for illustration. We can verify that the total number of parameters in this model is only $\mathcal{Z}_2 + 6 \times \mathcal{Z}_1$. In contrast, the number of parameters in the original naive model is $\mathcal{Z}_1 \times \mathcal{Z}_2$ which is much larger ($\mathcal{Z}_1$ and $\mathcal{Z}_2$ are usually $64^2$). The notable reduction of the learnable parameters is critical to improve the adaptation capability of \emph{MetaFuse}. Please refer to the Spatial Transformer Network \cite{jaderberg2015spatial} for more details about the implementation of $\Tau$.

\begin{figure}
	\centering
	\includegraphics[width=3.2in]{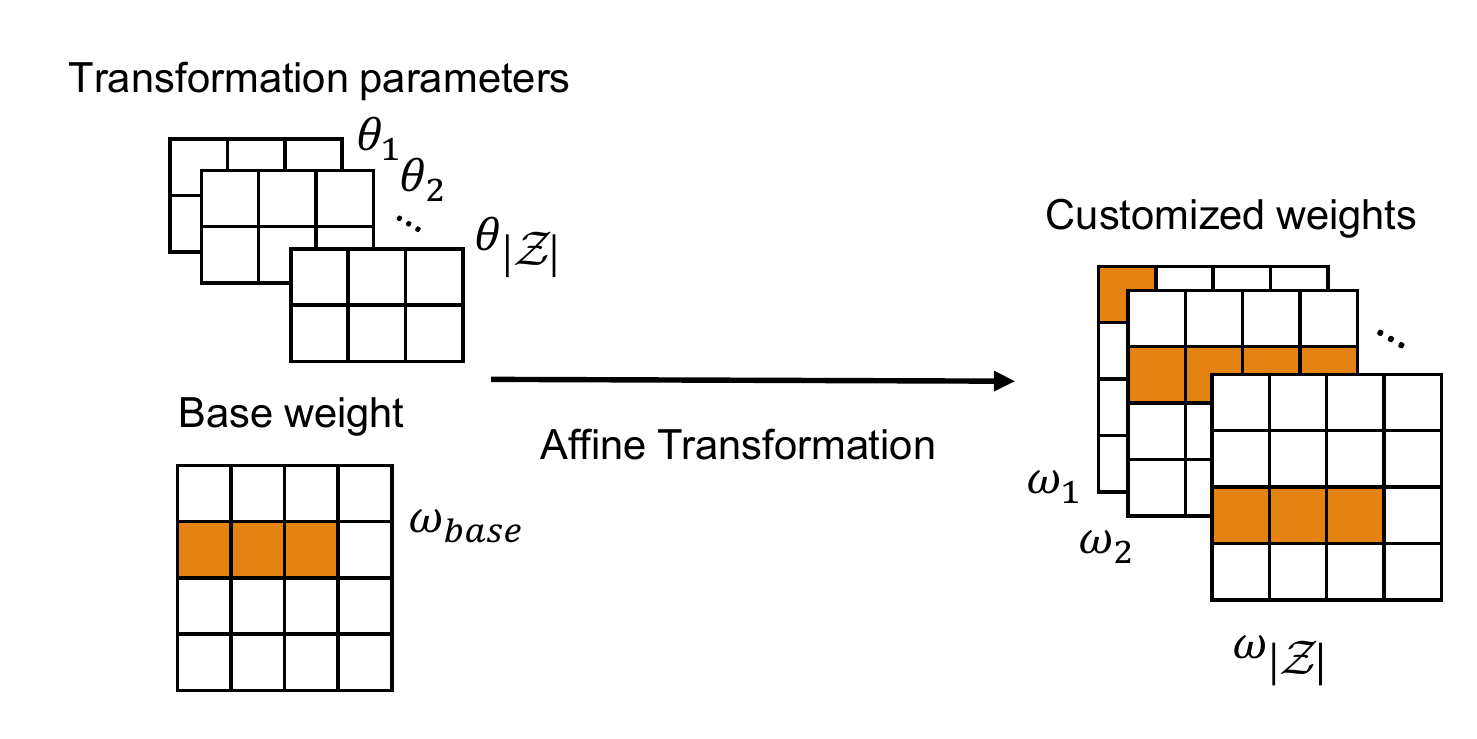}
	\caption{Applying different affine transformations $\Tau^{\theta_i}(\cdot)$ to the generic base weight $\omega^{\text{base}}$ to obtain the customized fusion weight $\omega_i$ for each pixel in view one. }
	\label{fig:base_weight}
\end{figure}

With sufficient image and pose annotations from a pair of cameras, we can directly learn the generic model $\bm{\omega^{\text{base}}}$ and the affine transformation parameters $\theta_i$ for every pixel by minimizing the following loss function:
\begin{equation}
\mathcal{L}_\mathcal{D_{\text{Tr}}}(\bm{\omega^{\text{base}}},\theta) = \frac{1}{|\mathcal{D_{\text{Tr}}}|} \sum_{  \mathcal{F},\mathcal{F}_{gt}\in \mathcal{D_{\text{Tr}}} }\text{MSE}(f_{[\bm{\omega^{\text{base}}};\theta]}(\mathcal{F}),\mathcal{F}_{gt}),
\label{eq:loss_t}
\end{equation}
where $\mathcal{F}$ are the initially estimated heatmaps (before fusion), and $f_{[\bm{\omega^{\text{base}}};\theta]}$ denotes the fusion function with parameters $\bm{\omega^{\text{base}}}$ and $\theta$. See Eq.(\ref{eq:update_rule}) and Eq.(\ref{eq:affine_rule}) for how we construct the fusion function. $\mathcal{F}_{gt}$ denotes the ground-truth pose heatmaps. Intuitively, we optimize $\bm{\omega^{\text{base}}}$ and $\theta$ such as to minimize the difference between the fused heatmaps and the ground truth heatmaps. We learn different $\theta$s for different pixels and camera pairs. It is also worth noting that both $\theta$ and $\bm{\omega^{\text{base}}}$ are global variables which do not depend on images. The loss function can be simply minimized by stochastic gradient descent. However, the model trained this way cannot generalize to new cameras with sufficient accuracy when only a few labeled data are available.

\subsection{Learning MetaFuse}
\label{sec:metalearning}
We now describe how we learn \emph{MetaFuse} including the generic model (\ie $\bm{\omega^{\text{base}}}$ and the initializations of $\theta$) from a large number of cameras so that the learned fusion model can rapidly adapt to new cameras using small data. The algorithm is inspired by a meta-learning algorithm proposed in \cite{finn2017model}. We describe the main steps for learning MetaFuse in the following subsections.

\begin{figure}
	\centering
	\includegraphics[width=1\linewidth]{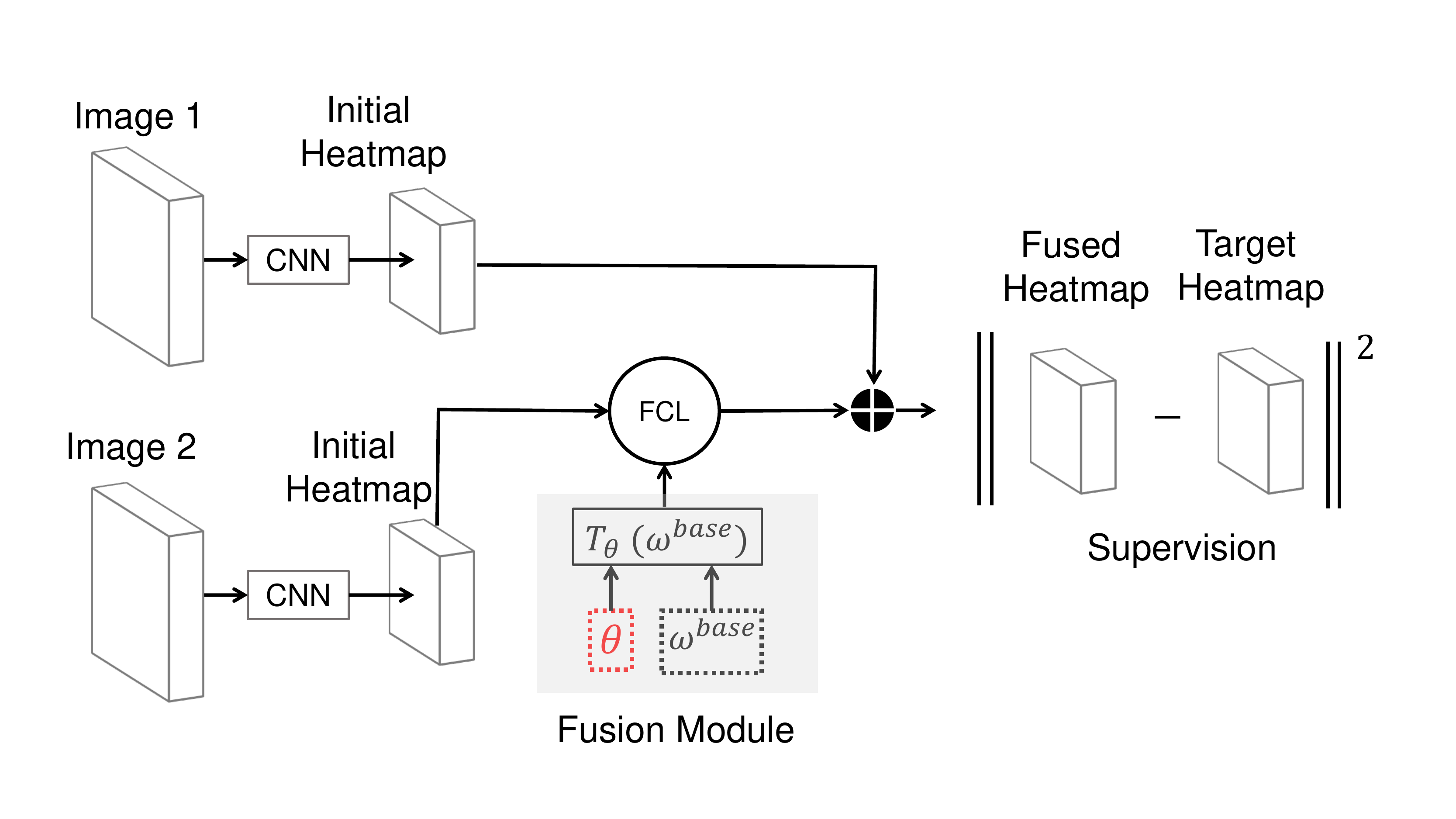}
	\caption{The pipeline for training \emph{MetaFuse}. In the first step, we pre-train the backbone network before fusion on all training images by regular gradient descent. In the second step, we fix the backbone parameters and meta-train $\bm{\omega^{\text{base}}}$ and $\theta$. In the testing stage, for a new camera configuration, we fix $\bm{\omega^{\text{base}}}$ and only finetune the transformation parameters $\theta$ based on small training data from the target camera. }
	\label{fig:framework}
\end{figure}

\paragraph{Warming Up} 
In the first step, we train the backbone network (\ie the layers before the fusion model) to speed up the subsequent meta-training process.  All images from the training dataset are used for training the backbone. The backbone parameters are directly optimized by minimizing the MSE loss between the initial heatmaps and the ground truth heatmaps. Note that the backbone network is only trained in this step, and will be fixed in the subsequent meta-training step to notably reduce the training time.

\paragraph{Meta-Training}
In this step, as shown in Figure \ref{fig:framework}, we learn the generic fusion model $\bm{\omega^{\text{base}}}$ and the initializations of $\theta$ by a meta-learning style algorithm. Generally speaking, the two parameters are sequentially updated by computing gradients over pairs of cameras (sampled from the dataset) which are referred to as tasks.

\emph{Task} is an important concept in meta-training. In particular, every task $\mathcal{T}_i$ is associated with a small dataset $\mathcal{D}_i$ which consists of a few images and ground truth $2$D pose heatmaps sampled from the same camera pair. For example, the camera pair $(\text{Cam}_1, \text{Cam}_2)$ is used in task $\mathcal{T}_1$ while the camera pair $(\text{Cam}_3, \text{Cam}_4)$ is used in in task $\mathcal{T}_2$.  We learn the fusion model from many of such different tasks so that it can get good results when adapted to a new task by only a few gradient updates. Let $\{\mathcal{T}_1, \mathcal{T}_2, \cdots, \mathcal{T}_N\}$ be
a number of tasks. Each $\mathcal{T}_i$ is associated with a dataset $\mathcal{D}_i$ consisting of data from a particular camera pair. Specifically, each $\mathcal{D}_i$ consists of two subsets: $\mathcal{D}_i^{train}$ and $\mathcal{D}_i^{test}$. As will be clarified later, both subsets are used for training. 

We follow the model-agnostic meta-learning framework \cite{finn2017model} to learn the optimal initializations for $\bm{\omega^{\text{base}}}$ and $\theta$. In the meta-training process, when adapted to a new task $\mathcal{T}_i$, the model parameters $\bm{\omega^{\text{base}}}$ and $\theta$ will become $\bm{\omega^{\text{base}}{'}}$ and $\theta'$, respectively. The core of meta-training is that we learn the optimal $\bm{\omega^{\text{base}}}$ and $\theta$ which will get a small loss on this task if it is updated based on the small dataset of the task. Specifically, $\bm{\omega^{\text{base}}{'}}$ and $\theta'$ can be computed by performing gradient descent on task $\mathcal{T}_i$
\begin{equation}
\theta' = \theta-\alpha \nabla_\theta\mathcal{L}_{\mathcal{D}^{\text{train}}_i}(\bm{\omega^{\text{base}}},\theta)
\label{eq:theta_update}
\end{equation}

\begin{equation}
\bm{\omega^{\text{base}\prime}} = \bm{\omega^{\text{base}}}-\alpha \nabla_{\bm{\omega^{\text{base}}}}\mathcal{L}_{\mathcal{D}^{\text{train}}_i}(\bm{\omega^{\text{base}}},\theta).
\label{eq:omega_update}
\end{equation}

The learning rate $\alpha$ is a hyper-parameter. It is worth noting that we do not actually update the model parameters according to the above equations. $\bm{\omega^{\text{base}\prime}}$ and $\theta'$ are the intermediate variables as will be clarified later. The core idea of meta learning is to learn $\bm{\omega^{\text{base}}}$ and $\theta$ such that after applying the above gradient update, the loss for the current task (evaluated on $\mathcal{D}_i^{test}$) is minimized.
The model parameters are trained by optimizing for the performance of $\mathcal{L}_{\mathcal{D}^{\text{test}}_i}(\bm{\omega^{\text{base}\prime}},\theta')$ with respect to $\bm{\omega_{\text{base}}}$ and $\theta$, respectively, across all tasks. 
Note that, $\bm{\omega^{\text{base}}{'}}$ and $\theta'$ are related to the initial parameters $\bm{\omega_{\text{base}}}$ and $\theta$ because of Eq.(\ref{eq:theta_update}) and Eq.(\ref{eq:omega_update}). More formally, the meta-objective is as follows:
\begin{equation}
\mathop{min}\limits_{\bm{\omega^{\text{base}}},\theta} \mathcal{L}_{\mathcal{D}^{\text{test}}_i}(\bm{\omega^{\text{base}\prime}},\theta')
\label{eq:meta_update}
\end{equation}

The optimization is performed over the parameters  $\bm{\omega_{\text{base}}}$ and $\theta$, whereas the objective is computed using the updated model parameters $\bm{\omega^{\text{base}\prime}}$ and $\theta'$. In effect, our method aims to optimize the model parameters such that one or a small number of gradient steps on a new task will produce maximally effective behavior on that task. We repeat the above steps iteratively on each task $\mathcal{D}_i \in \{\mathcal{D}_1, \mathcal{D}_2, \cdots, \mathcal{D}_N\}$. Recall that each $\mathcal{D}_i$ corresponds to a different camera configuration. So it actually learns a generic $\bm{\omega^{\text{base}}}$ and $\theta$ which can be adapted to many camera configurations with the gradients computed on small data. The $\bm{\omega^{\text{base}}}$  will be fixed after this meta-training stage.

\subsection{Finetuning MetaFuse}
For a completely new camera configuration, we adapt the meta-trained model by finetuning $\theta$. This is realized by directly computing the gradients to $\theta$ on a small number of labeled training data. Due to the lack of training data, the generic model $\bm{\omega^{\text{base}}}$ which is shared by all camera configurations will not be updated. The number of learnable parameters in this step is $6 \times H \times W$ which is only several thousands in practice.

\begin{figure*}
	\centering
	\includegraphics[width=6.2in]{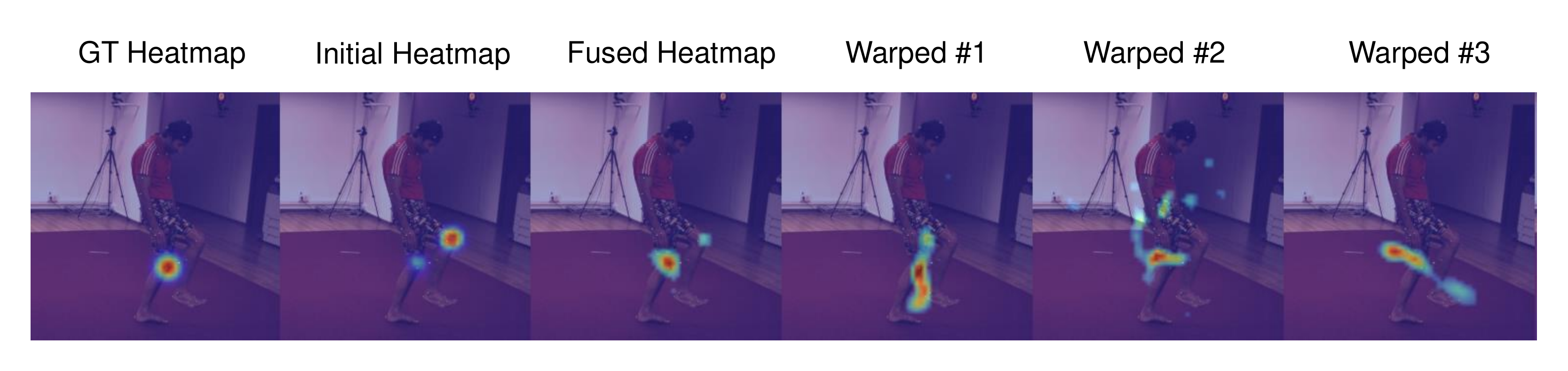}
	\caption{Heatmaps estimated by \emph{MetaFuse}. The first figure shows the ground truth heatmap of left knee. The second shows the initially detected heatmap. The highest response is at the wrong location. The third image shows the fused heatmap which correctly localizes the left knee. The rest images show the heatmaps warped from the other three views.}
	\label{fig:heatmap}
\end{figure*}

\begin{table*}[h]
	\caption{Description of the baselines.}
	\label{table:baseline}
	\centering
	\begin{tabularx}{0.95\linewidth}{l|X}
		\toprule
		\textbf{Names} & \textbf{Description} \\
		\midrule
		\emph{No-Fusion} & This is a simple baseline which does not perform multi-view fusion. It is equivalent to estimating poses in each camera view independently. This approach has the maximum flexibility since it can be directly applied to new environments without adaptation.  \\
		\midrule
		\emph{NaiveFuse}\textsuperscript{full} & This baseline directly trains the NaiveFuse model using all images from the target camera configuration. This can be regarded as an upper bound for \emph{MetaFuse} when there are sufficient training images. This approach has the LEAST flexibility because it requires to label a large number of images from each target camera configuration.\\
		\midrule
		\emph{NaiveFuse}\textsuperscript{$K$} & This baseline pre-trains the \emph{NaiveFuse} model on the Panoptic dataset (using four selected cameras) by regular stochastic gradient descent. Then it finetunes the pre-trained model on $K$ images from the target camera configuration. The approach is flexible when $K$ is small.  \\
		\midrule
		\emph{AffineFuse}\textsuperscript{$K$} & This baseline first pre-trains our factorized fusion model according to the description in Section \ref{sec:affinefusion} on the Panoptic dataset (using four selected cameras) by regular stochastic gradient descent. Then it finetunes the model on the target camera configuration. \\
		\midrule
		\emph{MetaFuse}\textsuperscript{K} & It finetunes the meta-learned model on $K$ images of the target cameras. It differs from \emph{AffineFuse}\textsuperscript{$K$} in that it uses the meta-learning style algorithm to pre-train the model.\\
		\bottomrule
	\end{tabularx}
\end{table*}

\section{Experiments} \label{sec:experiments}
\subsection{Datasets, Metrics and Details}
\paragraph{CMU Panoptic Dataset}
This dataset \cite{joo2015panoptic} provides images captured by a large number of synchronized cameras. We follow the convention in \cite{xiang2019monocular} to split the training and testing data. We select $20$ cameras (\ie $380$ ordered camera pairs) from the training set to pre-train \emph{MetaFuse}. Note that we only perform pre-training on this large dataset, and directly finetune the learned model on each target dataset to get a customized multi-view fusion based $2$D pose estimator. For the sake of evaluation on this dataset, we select six from the rest of the cameras. We run multiple trials and report average results to reduce the randomness caused by camera selections. In each trial, four cameras are chosen from the six for multi-view fusion. 

\paragraph{H36M Dataset}
This dataset \cite{ionescu2014human3} provides synchronized four-view images. We use subjects $1,5,6,7,8$ for finetuning the pre-trained model, and use subjects $9,11$ for testing purpose. It is worth noting that the camera placement is slightly different for each of the seven subjects. 

\paragraph{Total Capture Dataset}
In this dataset \cite{trumble2017total}, there are five subjects performing four actions including Roaming(\textbf{R}), Walking(\textbf{W}), Acting(\textbf{A}) and Freestyle(\textbf{FS}) with each repeating $3$ times.
We use Roaming 1,2,3, Walking 1,3, Freestyle 1,2 and Acting 1,2 of \emph{Subjects 1,2,3} for finetuning the pre-trained model. We test on Walking 2, Freestyle 3 and Acting 3 of all subjects. 

\paragraph{Metrics}
The $2$D pose accuracy is measured by Joint Detection Rate (JDR). If the distance between the estimated and the ground-truth joint location is smaller than a threshold, we regard this joint as successfully detected. The threshold is set to be half of the head size as in \cite{andriluka20142D}. JDR is computed as the percentage of the successfully detected joints. The $3$D pose estimation accuracy is measured by the Mean Per Joint Position Error (MPJPE) between the ground-truth $3$D pose and the estimation.
We do not align the estimated $3$D poses to the ground truth as in \cite{martinez2017simple,tome2018rethinking}.

\paragraph{Complexity} 
The warming up step takes about $30$ hours on a single 2080Ti GPU. The meta-training stage takes about $5$ hours. This stage is fast because we use the pre-computed heatmaps. The meta-testing (finetuning) stage takes about $7$ minutes. Note that, in real deployment, only meta-testing needs to be performed for a new environment which is very fast. In testing, it takes about $0.015$ seconds to estimate a $2$D pose from a single image.

\paragraph{Implementation Details}
We use a recent $2$D pose estimator \cite{simplebaselines} as the basic network to estimate the initial heatmaps. The ResNet50 \cite{he2016deep} is used as its backbone. The input image size is $256 \times 256$ and the resolution of the heatmap is 64 $\times$ 64. 
In general, using stronger 2D pose estimators can further improve the final 2D and 3D estimation results but that is beyond the scope of this work. We apply softmax with temperature $T=0.2$ to every channel of the fused heatmap to highlight the maximum response.

The Adam \cite{kingma2014adam} optimizer is used in all phases. We train the backbone network for $30$ epochs 
on the target dataset
in the warming up stage. Note that we do not train the fusion model in this step. The learning rate is initially set to be $1e^{-3}$, and drops to $1e^{-4}$ at 15 epochs and $1e^{-5}$ at 25 epochs, respectively. In meta-training and meta-testing, the learning rates are set to be $1e^{-3}$ and $5e^{-3}$, respectively. We evaluate our approach by comparing it to the five related baselines, which are detailed in Table \ref{table:baseline}.

\begin{figure*}[!htbp]
	\centering
	\includegraphics[width=7in]{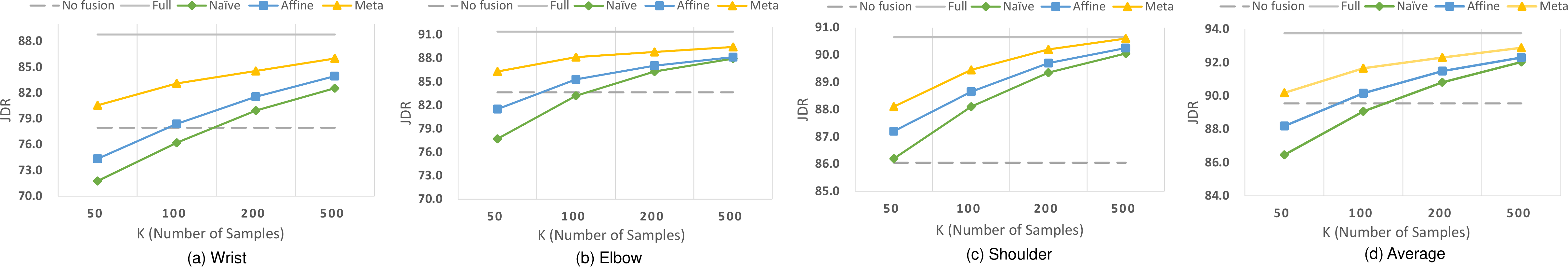}
	\caption{The $2$D joint detection rates of different methods on the H36M dataset. The x-axis represents the number of samples for finetuning the fusion model. The y-axis denotes the JDR. We show the average JDR over all joints, as well as the JDRs for several typical joints. The method of ``full'' denotes \emph{NaiveFuse}\textsuperscript{full} which can be regarded as an upper bound for all fusion methods.}
	\label{fig:2d_result}
\end{figure*}

\subsection{Results on the H36M Dataset}
\paragraph{2D Results}

The joint detection rates (JDR) of the baselines and our approach are shown in Figure \ref{fig:2d_result}. We present the average JDR over all joints, as well as the JDRs of several typical joints. We can see that the JDR of \emph{No-Fusion} (the grey dashed line) is lower than our \emph{MetaFuse} model regardless of the number of images used for finetuning the fusion model. This validates the importance of multi-view fusion. The improvement is most significant for the wrist and elbow joints because they are frequently occluded by human body in this dataset. 

\emph{NaiveFuse}\textsuperscript{full} (the grey solid line) gets the highest JDR because it uses all training data from the H36M dataset. However, when we use fewer data, the performance drops significantly (the green line). In particular, \emph{NaiveFuse}\textsuperscript{$50$} even gets worse results than \emph{No-Fusion}. This is because small training data usually leads to over-fitting for large models. We attempted to use several regularization methods including $l_2$, $l_1$ and $L_{2,1}$ (group sparsity) on $\bm{\omega}$ to alleviate the over-fitting problem of \emph{NaiveFuse}. But none of them gets better performance than vanilla \emph{NaiveFuse}. It means that the use of geometric priors in \emph{MetaFuse} is more effective than the regularization techniques.

Our proposed \emph{AffineFuse}\textsuperscript{$K$}, which has fewer parameters than \emph{NaiveFuse}\textsuperscript{$K$}, also gets better result when the number of training data is small (the blue line). However, it is still worse than \emph{MetaFuse}\textsuperscript{$K$}. This is because the model is not pre-trained on many cameras to improve its adaptation performance by our meta-learning-style algorithm which limits its performance on the H36M dataset.

Our approach \emph{MetaFuse}\textsuperscript{$K$} outperforms all baselines. In particular, it outperforms \emph{No-Fusion} when only $50$ training examples from the H36M dataset are used. Increasing this number consistently improves the performance. The result of \emph{MetaFuse}\textsuperscript{$500$} is already similar to that of \emph{NaiveFuse}\textsuperscript{full} which is trained on more than $80$K images. 

We also evaluate a variant of \emph{NaiveFuse} which is learned by the meta-learning algorithm. The average JDR are $87.7\%$ and $89.3\%$ when 50 and 100 examples are used,  which are much worse than \emph{MetaFuse}. The results validate the importance of the geometry inspired decomposition.

\begin{table}[ht]
	\center
	\caption{The $3$D MPJPE errors obtained by the state-of-the-art methods on the H36M dataset. \emph{MetaFuse} uses the pictorial model for estimating 3D poses. ``Full H36M Training'' means whether we use the full H36M dataset for adaptation or training. }
	\label{table:state_of_art_3D}
	\begin{tabular}{l  c c}
		\toprule
		Methods & Full H36M Training & MPJPE \\
		\toprule
		PVH-TSP \cite{trumble2017total} & \cmark & $87.3$mm\\
		Pavlakos \cite{PavlakosZDD17} &  \cmark & $56.9$mm\\
		Tome \cite{tome2018rethinking} &  \cmark & $52.8$mm \\
		Liang \cite{liang2019shape}&  \cmark & $45.1$mm \\
		CrossView \cite{qiu2019cross} & \cmark & $26.2$mm \\
		Volume \cite{iskakov2019learnable} & \cmark & $20.8$mm \\
		\midrule
		CrossView \cite{qiu2019cross} & \xmark & $43.0$mm \\
		Volume \cite{iskakov2019learnable} & \xmark & $34.0$mm \\
		\emph{MetaFuse}\textsuperscript{$50$} & \xmark &  32.7mm \\
		\emph{MetaFuse}\textsuperscript{$100$} & \xmark &  31.3mm \\
		\emph{MetaFuse}\textsuperscript{$500$} & \xmark &  \textbf{29.3}mm \\
		\toprule
	\end{tabular}
\end{table}

\paragraph{Examples}
Figure \ref{fig:heatmap} explains how \emph{MetaFuse} improves the $2$D pose estimation accuracy. The target joint is the left knee in this example. But the estimated heatmap (before fusion) has the highest response at the incorrect location (near right knee). By leveraging the heatmaps from the other three views, it accurately localizes the left knee joint. The last three images show the warped heatmaps from the other three views. We can see the high response pixels approximately form a line (the epipolar line) in each view.

\begin{figure*}[!htbp]
	\centering
	\includegraphics[width=6.0in]{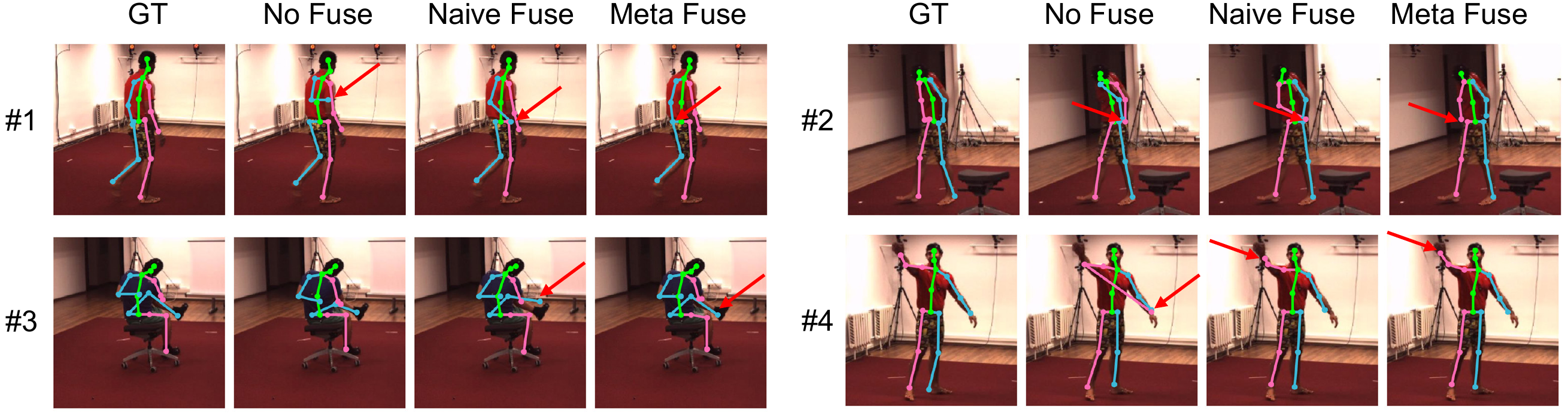}
	\caption{Four groups of sample $2$D poses estimated by different methods. Each group has $1$x$4$ sub-figures which correspond to the ground truth(GT) and three methods, respectively. The pink and cyan joints belong to the right and left body parts, respectively. The red arrows highlight the joints whose estimations are different for the three methods.}
	\label{fig:samples}
\end{figure*}

We visualize some typical poses estimated by the baselines and our approach in Figure \ref{fig:samples}. First, we can see that when occlusion occurs, \emph{No-Fusion} usually gets inaccurate $2$D locations for the occluded joints. For instance, in the first example, the left wrist joint is localized at a wrong location.  \emph{NaiveFuse} and \emph{MetaFuse} both help to localize the wrist joint in this example, and \emph{MetaFuse} is more accurate. However, in some cases, \emph{NaiveFuse} may get surprisingly bad results as shown in the third example. The left ankle joint is localized at a weird location even though it is visible. The main reason for this abnormal phenomenon is that the \emph{NaiveFuse} model learned from few data lacks generalization capability. \emph{MetaFuse} approach gets consistently better results than the two baseline methods.

\begin{table*}[!htbp]
	\center
	\caption{$3$D pose estimation errors MPJPE ($mm$) of different methods on the Total Capture dataset.}
	\label{table:totalcapture}
		\begin{tabular}{l c c  c c c c c c c}
			\toprule
			Approach  & \multicolumn{3}{c}{Subjects(S1,2,3)} & \multicolumn{3}{c}{Subjects(S4,5)} & Mean \\
			& Walking2 & Acting3  & FreeStyle3 & Walking2 & Acting3 & FreeStyle3 & \\
			\hline
			Tri-CPM \cite{wei2016convolutional}  & 79.0 &  106.5&  112.1& 79.0 & 73.7 & 149.3 & 99.8\\
			PVH \cite{trumble2017total}  & 48.3 & 94.3 & 122.3 & 84.3 & 154.5 & 168.5 & 107.3 \\
			IMUPVH \cite{trumble2017total} & 30.0 & 49.0 & 90.6 & 36.0 & 109.2 &  112.1 & 70.0\\
			LSTM-AE \cite{trumble2018deep} & \textbf{13.0} & \textbf{23.0} & 47.0 & \textbf{21.8} & 40.9 & 68.5 & 34.1 \\
			\hline
			\emph{No-Fusion} & 28.1 & 30.5 & 42.9 & 45.6 & 46.3 & 74.3 & 41.2 \\
			\emph{MetaFuse}\textsuperscript{$500$}  & 21.7 & 23.3 & \textbf{32.1} & 35.2 & \textbf{34.9} & \textbf{57.4} & \textbf{32.4} \\
			\toprule
		\end{tabular}
\end{table*}

\paragraph{3D Results}
We estimate $3$D pose from multi-view $2$D poses by a pictorial structure model \cite{qiu2019cross}. The results on the H36M dataset are shown in Table \ref{table:state_of_art_3D}. Our \emph{MetaFuse} trained on only $50$ examples decreases the error to $32.7$mm. Adding more training data consistently decreases the error. Note that some approaches in the table which use the full H36M dataset for training are not comparable to our approach. 

\subsection{Results on Total Capture}
The results are shown in Table \ref{table:totalcapture}. We can see that  \emph{MetaFuse} outperforms \emph{No-Fusion} by a large margin on all categories which validates its strong generalization power. In addition, our approach also outperforms the state-of-the-art ones including a recent work which utilizes temporal information \cite{trumble2018deep}. We notice that LSTM-AE \cite{trumble2018deep} outperforms our approach on the ``Walking2'' action. This is mainly because LSTM-AE uses temporal information which is very effective for this ``Walking2'' action. We conduct a simple proof-of-concept experiment where we apply the Savitzky-Golay filter \cite{schafer2011savitzky} to smooth the $3$D poses obtained by our approach. We find the average $3$D error for the ``Walking'' action of our approach decreases by about $5$mm. The result of our approach is obtained when \emph{MetaFuse} is finetuned on only $500$ images. In contrast, the state-of-the-art methods train their models on the whole dataset.

\subsection{Results on Panoptic Dataset}
We also conduct experiments on the Panoptic dataset. Note that the cameras selected for testing are different from those selected for pre-training.
The $3$D error of the \emph{No-Fusion} baseline is $40.47$mm. Our \emph{MetaFuse} approach gets a smaller error of $37.27$mm when only $50$ examples are used for meta-testing. This number further decreases to $31.78$mm when we use $200$ examples. In contrast. the errors for the \emph{NaiveFuse} approach are $43.39$mm and $35.60$mm when the training data number is $50$ and $200$, respectively. The results validate that our proposed fusion model can achieve consistently good results on the three large scale datasets.

\section{Conclusion}
We present a multi-view feature fusion approach which can be trained on as few as $100$ images for a new testing environment. It is very flexible in terms of that it can be integrated with any of the existing $2$D pose estimation networks, and it can be adapted to any environment with any camera configuration. The approach achieves the state-of-the-art results on three benchmark datasets. In our future work, we will explore the possibility to apply the fusion model to other tasks such as semantic segmentation. Besides, we can leverage synthetic data of massive cameras to further improve the generalization ability of model.

\paragraph{Acknowledgement}
This work was supported in part by MOST-2018AAA0102004, NSFC-61625201, 61527804 and DFG TRR169 / NSFC Major International Collaboration Project "Crossmodal Learning".

\small

\bibliographystyle{ieee_fullname}
\bibliography{egbib}

\begin{thebibliography}{10}\itemsep=-1pt

\bibitem{amin2013multi}
Sikandar Amin, Mykhaylo Andriluka, Marcus Rohrbach, and Bernt Schiele.
\newblock Multi-view pictorial structures for {3D} human pose estimation.
\newblock In {\em BMVC}, 2013.

\bibitem{andriluka20142D}
Mykhaylo Andriluka, Leonid Pishchulin, Peter Gehler, and Bernt Schiele.
\newblock {2D} human pose estimation: New benchmark and state of the art
  analysis.
\newblock In {\em CVPR}, pages 3686--3693, 2014.

\bibitem{andrychowicz2016learning}
Marcin Andrychowicz, Misha Denil, Sergio Gomez, Matthew~W Hoffman, David Pfau,
  Tom Schaul, Brendan Shillingford, and Nando De~Freitas.
\newblock Learning to learn by gradient descent by gradient descent.
\newblock In {\em NIPS}, pages 3981--3989, 2016.

\bibitem{belagiannis20143d}
Vasileios Belagiannis, Sikandar Amin, Mykhaylo Andriluka, Bernt Schiele, Nassir
  Navab, and Slobodan Ilic.
\newblock 3d pictorial structures for multiple human pose estimation.
\newblock In {\em CVPR}, pages 1669--1676, 2014.

\bibitem{bo2010twin}
Liefeng Bo and Cristian Sminchisescu.
\newblock Twin gaussian processes for structured prediction.
\newblock {\em IJCV}, 87(1-2):28, 2010.

\bibitem{burenius20133D}
Magnus Burenius, Josephine Sullivan, and Stefan Carlsson.
\newblock {3D} pictorial structures for multiple view articulated pose
  estimation.
\newblock In {\em CVPR}, pages 3618--3625, 2013.

\bibitem{Chen_2019_CVPR}
Xipeng Chen, Kwan-Yee Lin, Wentao Liu, Chen Qian, and Liang Lin.
\newblock Weakly-supervised discovery of geometry-aware representation for 3d
  human pose estimation.
\newblock In {\em The IEEE Conference on Computer Vision and Pattern
  Recognition (CVPR)}, June 2019.

\bibitem{dong2019fast}
Junting Dong, Wen Jiang, Qixing Huang, Hujun Bao, and Xiaowei Zhou.
\newblock Fast and robust multi-person 3d pose estimation from multiple views.
\newblock In {\em CVPR}, pages 7792--7801, 2019.

\bibitem{duan2016rl}
Yan Duan, John Schulman, Xi Chen, Peter~L Bartlett, Ilya Sutskever, and Pieter
  Abbeel.
\newblock Rl$^{2}$: Fast reinforcement learning via slow reinforcement
  learning.
\newblock {\em arXiv preprint arXiv:1611.02779}, 2016.

\bibitem{finn2017model}
Chelsea Finn, Pieter Abbeel, and Sergey Levine.
\newblock Model-agnostic meta-learning for fast adaptation of deep networks.
\newblock In {\em ICML}, pages 1126--1135. JMLR. org, 2017.

\bibitem{gall2010optimization}
Juergen Gall, Bodo Rosenhahn, Thomas Brox, and Hans-Peter Seidel.
\newblock Optimization and filtering for human motion capture.
\newblock {\em IJCV}, 87(1-2):75, 2010.

\bibitem{hartley2003multiple}
Richard Hartley and Andrew Zisserman.
\newblock {\em Multiple view geometry in computer vision}.
\newblock Cambridge university press, 2003.

\bibitem{he2016deep}
Kaiming He, Xiangyu Zhang, Shaoqing Ren, and Jian Sun.
\newblock Deep residual learning for image recognition.
\newblock In {\em CVPR}, pages 770--778, 2016.

\bibitem{ionescu2014human3}
Catalin Ionescu, Dragos Papava, Vlad Olaru, and Cristian Sminchisescu.
\newblock Human3. 6m: Large scale datasets and predictive methods for {3D}
  human sensing in natural environments.
\newblock {\em T-PAMI}, pages 1325--1339, 2014.

\bibitem{iskakov2019learnable}
Karim Iskakov, Egor Burkov, Victor Lempitsky, and Yury Malkov.
\newblock Learnable triangulation of human pose.
\newblock In {\em ICCV}, 2019.

\bibitem{jaderberg2015spatial}
Max Jaderberg, Karen Simonyan, Andrew Zisserman, et~al.
\newblock Spatial transformer networks.
\newblock In {\em NIPS}, pages 2017--2025, 2015.

\bibitem{joo2015panoptic}
Hanbyul Joo, Hao Liu, Lei Tan, Lin Gui, Bart Nabbe, Iain Matthews, Takeo
  Kanade, Shohei Nobuhara, and Yaser Sheikh.
\newblock Panoptic studio: A massively multiview system for social motion
  capture.
\newblock In {\em ICCV}, pages 3334--3342, 2015.

\bibitem{kingma2014adam}
Diederik~P Kingma and Jimmy Ba.
\newblock Adam: A method for stochastic optimization.
\newblock In {\em ICLR}, 2015.

\bibitem{lee2018gradient}
Yoonho Lee and Seungjin Choi.
\newblock Gradient-based meta-learning with learned layerwise metric and
  subspace.
\newblock In {\em ICML}, pages 2933--2942, 2018.

\bibitem{liang2019shape}
Junbang Liang and Ming~C Lin.
\newblock Shape-aware human pose and shape reconstruction using multi-view
  images.
\newblock In {\em Proceedings of the IEEE International Conference on Computer
  Vision}, pages 4352--4362, 2019.

\bibitem{liu2011markerless}
Yebin Liu, Carsten Stoll, Juergen Gall, Hans-Peter Seidel, and Christian
  Theobalt.
\newblock Markerless motion capture of interacting characters using multi-view
  image segmentation.
\newblock In {\em CVPR}, pages 1249--1256. IEEE, 2011.

\bibitem{martinez2017simple}
Julieta Martinez, Rayat Hossain, Javier Romero, and James~J Little.
\newblock A simple yet effective baseline for {3D} human pose estimation.
\newblock In {\em ICCV}, page~5, 2017.

\bibitem{mishra2018a}
Nikhil Mishra, Mostafa Rohaninejad, Xi Chen, and Pieter Abbeel.
\newblock A simple neural attentive meta-learner.
\newblock In {\em ICLR}, 2018.

\bibitem{PavlakosZDD17}
Georgios Pavlakos, Xiaowei Zhou, Konstantinos~G. Derpanis, and Kostas
  Daniilidis.
\newblock Harvesting multiple views for marker-less {3D} human pose
  annotations.
\newblock In {\em CVPR}, pages 1253--1262, 2017.

\bibitem{qiu2019cross}
Haibo Qiu, Chunyu Wang, Jingdong Wang, Naiyan Wang, and Wenjun Zeng.
\newblock Cross view fusion for 3d human pose estimation.
\newblock In {\em ICCV}, pages 4342--4351, 2019.

\bibitem{ravi2016optimization}
Sachin Ravi and Hugo Larochelle.
\newblock Optimization as a model for few-shot learning.
\newblock In {\em ICLR}, 2017.

\bibitem{rhodin2018learning}
Helge Rhodin, J{\"o}rg Sp{\"o}rri, Isinsu Katircioglu, Victor Constantin,
  Fr{\'e}d{\'e}ric Meyer, Erich M{\"u}ller, Mathieu Salzmann, and Pascal Fua.
\newblock Learning monocular 3d human pose estimation from multi-view images.
\newblock In {\em CVPR}, pages 8437--8446, 2018.

\bibitem{santoro2016meta}
Adam Santoro, Sergey Bartunov, Matthew Botvinick, Daan Wierstra, and Timothy
  Lillicrap.
\newblock Meta-learning with memory-augmented neural networks.
\newblock In {\em ICML}, pages 1842--1850, 2016.

\bibitem{schafer2011savitzky}
Ronald~W Schafer et~al.
\newblock What is a savitzky-golay filter.
\newblock {\em IEEE Signal processing magazine}, 28(4):111--117, 2011.

\bibitem{snell2017prototypical}
Jake Snell, Kevin Swersky, and Richard Zemel.
\newblock Prototypical networks for few-shot learning.
\newblock In {\em NIPS}, pages 4077--4087, 2017.

\bibitem{Sun_2019_CVPR}
Qianru Sun, Yaoyao Liu, Tat-Seng Chua, and Bernt Schiele.
\newblock Meta-transfer learning for few-shot learning.
\newblock In {\em The IEEE Conference on Computer Vision and Pattern
  Recognition (CVPR)}, June 2019.

\bibitem{tome2018rethinking}
Denis Tome, Matteo Toso, Lourdes Agapito, and Chris Russell.
\newblock Rethinking pose in {3D}: Multi-stage refinement and recovery for
  markerless motion capture.
\newblock In {\em 3DV}, pages 474--483, 2018.

\bibitem{trumble2018deep}
Matthew Trumble, Andrew Gilbert, Adrian Hilton, and John Collomosse.
\newblock Deep autoencoder for combined human pose estimation and body model
  upscaling.
\newblock In {\em ECCV}, pages 784--800, 2018.

\bibitem{trumble2017total}
Matthew Trumble, Andrew Gilbert, Charles Malleson, Adrian Hilton, and John
  Collomosse.
\newblock Total capture: {3D} human pose estimation fusing video and inertial
  sensors.
\newblock In {\em BMVC}, pages 1--13, 2017.

\bibitem{vilalta2002perspective}
Ricardo Vilalta and Youssef Drissi.
\newblock A perspective view and survey of meta-learning.
\newblock {\em Artificial intelligence review}, 18(2):77--95, 2002.

\bibitem{wei2016convolutional}
Shih-En Wei, Varun Ramakrishna, Takeo Kanade, and Yaser Sheikh.
\newblock Convolutional pose machines.
\newblock In {\em CVPR}, pages 4724--4732, 2016.

\bibitem{xiang2019monocular}
Donglai Xiang, Hanbyul Joo, and Yaser Sheikh.
\newblock Monocular total capture: Posing face, body, and hands in the wild.
\newblock In {\em CVPR}, 2019.

\bibitem{simplebaselines}
Bin Xiao, Haiping Wu, and Yichen Wei.
\newblock Simple baselines for human pose estimation and tracking.
\newblock In {\em ECCV}, pages 466--481, 2018.

\bibitem{jafarian2018monet}
Yuan Yao, Yasamin Jafarian, and Hyun~Soo Park.
\newblock Monet: Multiview semi-supervised keypoint detection via epipolar
  divergence.
\newblock In {\em ICCV}, pages 753--762, 2019.

\end{thebibliography}

\clearpage

\begin{center}
	\textbf{\Large Supplementary Material}
\end{center}

\section{Additional Results on Panoptic Dataset}

\begin{table}[!htbp]
	\caption{$2$D pose estimation accuracy on the Panoptic dataset. 
		The second column represents the number of samples for finetuning the  pre-trained model. We report results for three joints and also the average result over all joints.}
	\label{table:2dpose}
	\centering
	{\small
		\begin{tabular}{lccccc}
			
			\toprule
			Methods  & Samples  & Shld. & Knee.  & Ankle.  & Avg    \\
			\midrule
			\emph{No-Fusion}   & -- & 89.9   & 79.8 & 89.7 & 88.9 \\
			\midrule
			\emph{NaiveFuse}\textsuperscript{$50$}     & \multirow{2}{*}{50}  & 88.1   & 82.3 & 87.6 & 85.1 \\
			\emph{MetaFuse}\textsuperscript{$50$}     &         & 91.2   & 85.7 & 90.8 & \textbf{88.9} \\
			\midrule
			\emph{NaiveFuse}\textsuperscript{$100$}    & \multirow{2}{*}{100}  & 88.8   & 83.7 & 87.6 & 86.5 \\
			\emph{MetaFuse}\textsuperscript{$100$}    &         & 91.4   & 86.3 & 90.9 & \textbf{89.5} \\
			\midrule
			\emph{NaiveFuse}\textsuperscript{$200$}     & \multirow{2}{*}{200} & 90.9   & 85.1 & 88.1 & 88.3 \\
			\emph{MetaFuse}\textsuperscript{$200$}    &     & 92.2   & 86.9 & 91.6 & \textbf{90.6} \\
			\midrule
			\emph{NaiveFuse}\textsuperscript{$500$}    & \multirow{2}{*}{500}     & 91.6   & 85.8 & 89.9 & 90.2 \\
			\emph{MetaFuse}\textsuperscript{$500$}    &         & 93.2   & 88.0 & 91.5 & \textbf{91.2} \\
			
			\bottomrule
		\end{tabular}
	}
\end{table}

\begin{table}[!htbp]
	\centering
	\caption{The $3$D pose MPJPE errors obtained by the baseline
		methods on the Panoptic dataset.  }
	\label{table:panoptic_3D}
	\begin{tabular}{p{58pt} p{60pt}<{\centering} p{80pt}<{\centering}}
		\toprule
		Methods & Samples &  Average MPJPE \\
		\toprule
		\emph{No-Fusion}& -- &   $40.47$mm\\
		\midrule
		\emph{NaiveFuse}\textsuperscript{$50$}& \multirow{2}{*}{50}  &   $43.39$mm \\
		\emph{MetaFuse}\textsuperscript{$50$}  &  & \textbf{37.27}mm \\
		\midrule
		\emph{NaiveFuse}\textsuperscript{$100$}& \multirow{2}{*}{100} &   $42.58$mm \\
		\emph{MetaFuse}\textsuperscript{$100$} &  & \textbf{36.02}mm \\
		\midrule
		\emph{NaiveFuse}\textsuperscript{$200$}& \multirow{2}{*}{200}  &   $35.60$mm \\
		\emph{MetaFuse}\textsuperscript{$200$}  &  & \textbf{31.78}mm \\
		\midrule
		\emph{NaiveFuse}\textsuperscript{$500$}& \multirow{2}{*}{500}  &  $33.50$mm \\
		\emph{MetaFuse}\textsuperscript{$500$}  &  &\textbf{30.88}mm \\
		\toprule
	\end{tabular}
\end{table}

\newpage
\section{Algorithm of Meta-Training}

\begin{algorithm}[htb]
	\setstretch{1.2}
	\caption{ Meta-Training of \emph{MetaFuse}}
	\label{alg:Framwork}
	\begin{algorithmic}[1]
		
		\Require
		$\{\mathcal{T}_1, \mathcal{T}_2, \cdots, \mathcal{T}_N\}:$ Each $\mathcal{T}_i$ is associated with a 
		
		small dataset from a particular camera pair.

		$\alpha,\beta:$ Step size, hyper-parameters
		\Ensure
		$\theta, \omega^{base}:$  Pre-trained fusion model
		\State Randomly initialize $\theta, \omega^{base}$
		
		\For{each $\mathcal{T}_i\in \{\mathcal{T}_1, \mathcal{T}_2, \cdots, \mathcal{T}_N\}$}
		\State Sample $K$ images $\mathcal{D}^{train}_i$ from $\mathcal{T}_i$  
		\State Compute $\theta', \omega^{base\prime}$
		with gradient descent on $\mathcal{D}^{train}_i$

		$
		\theta' = \theta-\alpha \nabla_\theta\mathcal{L}_{\mathcal{D}^{\text{train}}_i}(\bm{\omega^{\text{base}}},\theta)
		\label{eq:theta_update}
		$
		
		$
		\bm{\omega^{\text{base}\prime}} = \bm{\omega^{\text{base}}}-\alpha \nabla_{\bm{\omega^{\text{base}}}}\mathcal{L}_{\mathcal{D}^{\text{train}}_i}(\bm{\omega^{\text{base}}},\theta)
		\label{eq:omega_update}
		$
		
		\State Sample other $K$ images $\mathcal{D}^{test}_i$  from $\mathcal{T}_i$  
		\State Update $\theta \gets \theta - \beta \nabla_\theta \mathcal{L}_{\mathcal{D}^{\text{test}}_i}(\bm{\omega^{\text{base}\prime}},\theta')$

		$\bm{\omega^{\text{base}}} \gets \bm{\omega^{\text{base}}} - \beta \nabla_{\omega^{base}} \mathcal{L}_{\mathcal{D}^{\text{test}}_i}(\bm{\omega^{\text{base}\prime}},\theta')$
		\EndFor

		\\
		\Return $\theta, \omega^{base}$
	\end{algorithmic}
\end{algorithm}

\end{document}